\documentclass[
]{ceurart}

\usepackage{listings}
\usepackage{tikz}
\usepackage{pgfplots}
\pgfplotsset{compat=1.17}
\usepgfplotslibrary{statistics}
\usepgfplotslibrary{fillbetween}
\pgfplotsset{
	boxplot/hide outliers/.code={
		\def\pgfplotsplothandlerboxplot@outlier{}%
	}
}

\usepackage{color,xcolor}
\definecolor{gnome_blue}{HTML}{3465A4}
\definecolor{gnome_purple}{HTML}{75507B}
\definecolor{gnome_red}{HTML}{CC0000}
\definecolor{gnome_green}{HTML}{73D216}
\definecolor{gnome_yellow}{HTML}{EDD400}
\definecolor{gnome_orange}{HTML}{F57900}
\definecolor{gnome_brown}{HTML}{C17D11}
\definecolor{gnome_grey}{HTML}{555753}

\begin{document}

\copyrightyear{2026}
\copyrightclause{Copyright for this paper by its authors.
  Use permitted under Creative Commons License Attribution 4.0
  International (CC BY 4.0).}

\conference{8th International Workshop on Robotics Software Engineering (RoSE'26),
  June 01, 2026, Vienna, Austria}

\title{ROS2 Connect: A new ROS2 over WAN Solution}


\author[1]{Daniel Schott}[%
email=daniel.schott@uni-wuerzburg.de,
orcid=0009-0005-1926-9549
]

\author[1]{Lakshminarasimhan Srinivasan}[%
email=srinivasan@informatik.uni-wuerzburg.de,
orcid=0009-0003-9648-5574
]

\author[1]{Christian Herrmann}[%
email=christian.herrmann@uni-wuerzburg.de
]

\author[1,2,3]{Andreas Nüchter}[%
email=andreas.nuechter@uni-wuerzburg.de,
orcid=0000-0003-3870-783X
]

\address[1]{Computer Science XVII - Robotics, Julius-Maximilians-Universität Würzburg, Germany}
\address[2]{ENSTA, U2IS (Visiting Chair), Institut Polytechnique de Paris, Palaiseau, France}
\address[3]{Zentrum für Telematik e.V., Würzburg, Germany}

\begin{abstract}
  The Robot Operating System 2 (ROS2) has become a widely adopted framework for the development of distributed robotic systems.
  However, its communication architecture, based on DDS and RTPS, relies on multicast discovery mechanisms that are typically
  unavailable in wide-area network (WAN) environments, making remote operation challenging. This work presents ROS2 Connect, a
  WebSocket-based communication framework that enables transparent and secure ROS2 interaction across routed networks without
  requiring modifications to network infrastructure or DDS configurations. The proposed client-server architecture supports
  bidirectional exchange of topics, services, actions, and system data while integrating authentication and access control mechanisms.
  Experimental evaluation over a real WAN connection demonstrates significantly lower latency, higher stability, and improved
  scalability compared to existing solutions, including DDS Router, rosbridge and Zenoh. Initial results show that ROS2 Connect 
  provides a reliable foundation for teleoperation and distributed robotics applications over wide-area networks.
\end{abstract}


\maketitle

\section{Introduction}

Mobile robots have become increasingly common in distributed settings such as teleoperation, cloud robotics, and multi-site systems,
where communication across wide-area networks (WANs) becomes essential.
In  this context, the Robot Operating System 2 (ROS2) has emerged as the de facto standard framework for the development, integration 
and coordination of heterogeneous robotic systems \cite{ros2}.
Despite its widespread adoption, ROS2 remains primarily designed for local-area network (LAN) deployments.
Its communication layer, based on the Data Distribution Service (DDS)  and the Real-Time Publish-Subscribe Protocol (RTPS),
fundamentally relies on IP multicast for participant discovery and data exchange \cite{ros2, dds, rtps}.
This design assumption assumes multicast-capable network infrastructure, which is commonly available in local-area
networks but typically not enabled across routed WAN or Internet environments, where multicast requires explicit network-layer
support \cite{multicast}.
Consequently, ROS2 deployments over WAN suffer from impaired automatic discovery, increased configuration complexity, and 
reduced scalability. In particular, DDS/RTPS discovery mechanisms, which depend on multicast-based peer discovery, introduce 
substantial challenges when operating across routed and heterogeneous networks.
As a result, enabling efficient and stable ROS2 communication over WAN remains a non-trivial and
largely unresolved problem.

\noindent
To address these limitations, we propose a WebSocket-based communication framework for ROS2.
Our approach consists of a client-server architecture implemented as a ROS2 package,
allowing transparent bidirectional exchange of topics, services, actions, tf2 transforms, and time synchronization data
between distributed systems.
In addition to transport-layer mediation, the framework incorporates authentication and authorization mechanisms.
This design enables controlled remote access to robotic systems over routed and heterogeneous network infrastructures, while 
preserving operational safety and system integrity.
Experimental results demonstrate that ROS2 Connect exhibits lower latency while maintaining consistently low variability in comparison
with existing solutions. The implementation is publicly available at \url{https://github.com/JMUWRobotics/ROS2-Connect}.

\section{Existing Solutions}

Existing solutions for enabling ROS2 communication across WANs are broadly categorized into network-layer 
approaches, DDS-level routing mechanisms, and alternative middleware frameworks.
Network-layer solutions typically rely on virtual private networks (VPNs) to extend local-area connectivity across routed 
infrastructures, allowing native DDS discovery and communication to operate transparently. However, VPN deployment is often 
impractical in centrally managed environments, such as organizational networks, where users lack the privileges required to establish 
custom network tunnels. Consequently, VPN-based approaches are not considered further in this work.
Middleware-level approaches modify DDS discovery behavior. Examples include vendor-specific mechanisms such as the centralized 
discovery server in eProsima's Fast DDS and static unicast peer configurations supported by Eclipse Foundation's Cyclone DDS 
\cite{fastdds,cyclonedds}. While these methods enable WAN communication, they introduce dependencies on specific DDS implementations 
and remain difficult to deploy across network address translation (NAT) boundaries, as these approaches rely on direct peer
addressability and are not designed for traversal or dynamic endpoint negotiation.
Another solution is eProsima's DDS Router, which forwards DDS/RTPS traffic over TCP/IP and supports relay-based NAT traversal 
\cite{ddsrouter}. Although this approach enables communication across complex network topologies, it introduces additional routing 
overhead and remains tightly coupled to DDS-specific infrastructure.
Alternative frameworks operate outside the DDS ecosystem. The rosbridge interface enables WebSocket-based communication by 
serializing ROS messages into JSON \cite{rosbridge}, but it is commonly used for web-client integration and requires additional 
application-layer handling, while JSON serialization introduces significant communication overhead.
Eclipse Zenoh provides a modern publish-subscribe protocol with an available ROS middleware (RMW) implementation and support for 
multiple transport mechanisms, including WebSockets and TCP/IP \cite{zenoh}. Similar to eProsima's DDS Router, it enables 
communication across complex network topologies and supports NAT traversal. However, Zenoh replaces DDS entirely and requires 
additional integration effort.

\noindent
Despite the availability of these approaches, achieving transparent, secure, and performant ROS2 communication across WANs
without requiring vendor-specific dependencies, network-layer modifications, or substantial integration effort remains
an open challenge. In particular, existing solutions either rely on infrastructure-level adaptations, introduce tight coupling
to specific DDS implementations, impose significant configuration overhead, or fail to provide integrated mechanisms for secure 
and controlled remote access.
To address these limitations, this work proposes a vendor-independent communication framework that enables native ROS2 interaction
across WAN environments without requiring modifications to underlying network infrastructure. The proposed approach leverages
a WebSocket-based client-server architecture to mediate ROS2 communication while preserving transparency at the application
level and providing integrated support for authentication and access restriction.

\section{ROS2 Connect}

ROS2 Connect is a communication framework designed to enable transparent and secure interaction between distributed ROS2 systems
across WANs. The framework addresses the limitations of multicast-dependent DDS discovery by introducing a 
transport-layer mediation mechanism that decouples ROS2 communication from the underlying network topology.
The proposed architecture is implemented as a ROS2 package that provides two complementary node types: a server node and a client
node. The server node is deployed on a publicly reachable host within the robot's local ROS2 environment and communicates with local
robotic components using the native DDS-based ROS2 middleware. The client node, running on a remote operator system,
establishes a persistent WebSocket connection to the server node across the WAN.
Through this architecture, ROS2 communication data (topics, services, actions) is transparently tunneled between server and
client nodes. The nodes bridge between DDS-based local communication and WebSocket-based wide-area transport, enabling
remote interaction without requiring multicast support, VPN infrastructure, or vendor-specific DDS configuration.
WebSockets are employed as the underlying transport protocol due to their compatibility with firewall and NAT environments,
and ability to provide reliable, bidirectional communication over standard TCP/IP networks. In addition, WebSocket-based
communication has previously been successfully applied in practical teleoperation systems for mobile robots, demonstrating
its suitability for latency-sensitive remote control scenarios \cite{laks}.

\subsection{Communication Model}

ROS2 Connect mediates communication between distributed ROS2 systems by explicitly managing the exchange of ROS interfaces
across a client-server WebSocket connection. Rather than relying on DDS multicast discovery, the framework employs a 
configuration-driven model in which both server and client nodes define the set of topics, services, actions, and their
associated QoS profiles to be relayed through ROS parameters. This approach ensures that only explicitly authorized interfaces 
are exposed across the wide-area link.
For topic-based communication, ROS2 Connect utilizes generic subscription and publication mechanisms provided by
\texttt{rclcpp::GenericSubscription} and \texttt{rclcpp::GenericPublisher}. These interfaces allow the framework to operate
independently of compile-time message definitions by handling serialized binary message data at runtime.
Upon initialization, each node creates generic subscriptions and publishers for the configured topics, enabling the transparent
forwarding of arbitrary ROS message types. For transport, a lightweight protocol header containing topic identifiers and
compression metadata is prepended to each message before transmission over the WebSocket connection. 
This enables optional per-topic data compression to reduce bandwidth usage while allowing efficient unpacking and republishing on 
the receiving side.
To minimize bandwidth consumption and reduce processing, data forwarding is subscription-aware. Messages are
transmitted only when a corresponding subscriber exists on the remote system, thereby limiting traffic to actively used
interfaces and reducing load on both the DDS middleware and the WebSocket transport layer.
In addition to generic topic forwarding, ROS2 Connect provides dedicated support for core ROS2 infrastructure mechanisms 
required for distributed operation. The framework transparently relays the transform tree by forwarding the standard
\texttt{/tf} and \texttt{/tf\_static} topics, ensuring consistent spatial relationships between robot and operator systems.
Furthermore, time synchronization is supported by propagating the robot's ROS time through the \texttt{/clock} topic.
Service and action interfaces are handled through a plugin-based mechanism due to the lack of fully generic runtime support
in \texttt{rclcpp}. Type-specific service and action client and server implementations are provided as ROS2 Connect plugins,
which are dynamically loaded using the ROS2 plugin infrastructure (pluginlib). This design preserves the generic architecture
of the framework while enabling transparent mirroring of selected services and actions between distributed systems.
Through this combination of explicit interface mediation, runtime-generic message handling, and subscription-aware routing,
ROS2 Connect replaces multicast-dependent discovery with a secure communication model suitable for
wide-area deployments.

\subsection{Security and Access Control}

ROS2 Connect incorporates authentication and access control mechanisms to enable secure remote interaction with robotic systems. 
Authentication is performed at the application layer following the establishment of a WebSocket connection. Each client provides 
a user-specific authentication token configured through ROS parameters. Upon connection, the server initially suppresses all data 
exchange and processes only authentication messages. After successful verification of the provided credentials, the connection 
transitions into operational mode and regular communication is permitted.
To maintain flexibility and application independence, the authentication procedure is implemented through a plugin-based 
architecture. This allows system integrators to define custom authentication strategies, such as key-based validation or external 
identity management integration, without modifying the core framework.
Authorization is enforced through the configuration-driven interface mediation model. All topics, services, and actions that are 
relayed across the wide-area connection must be explicitly specified in advance, including their associated message types. The 
server only forwards data for these predefined interfaces and rejects any attempts to access or publish to unspecified topics, services,
or actions. This approach ensures that clients cannot access arbitrary system data or inject unauthorized messages.
Together, these mechanisms integrate security directly into the communication framework, enabling controlled remote operation while 
preserving the transparency of standard ROS 2 interfaces.

\section{Evaluation}

Experiments were conducted over a real WAN connection between a residential client system and a university-hosted
server. The server was accessible through a publicly reachable Apache reverse proxy deployed within the university network
infrastructure. Network bandwidth measurements using \textit{iperf3} indicate an available uplink capacity of approximately
18 Mbit/s and a downlink capacity of 58 Mbit/s.
All experiments were performed using ROS2 Jazzy Jalisco on Ubuntu 24.04. To evaluate communication performance, round-trip times
(RTT) were measured for serialized ROS messages with payload sizes ranging from 12 B to 500 kB.
The selected range reflects common robotic workloads from small scalar sensor values and control commands to medium-sized
LiDAR scan data and larger payloads such as camera images.
For each configuration, 1000 independent measurements were recorded to obtain statistically robust averages.
Additional experiments evaluated performance under concurrent load by transmitting between one and ten parallel topics. For each
configuration, 1000 round-trip measurements were recorded for message sizes ranging from 12 B to 100 kB.
The proposed ROS2 Connect framework was compared against three existing approaches: eProsima's DDS Router, the rosbridge WebSocket
interface, and Eclipse Zenoh. All systems were configured to operate under identical network conditions.

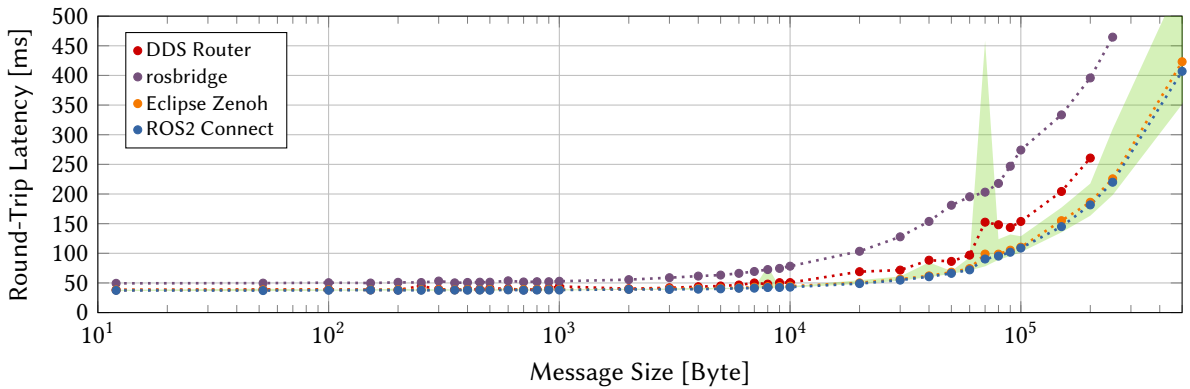
\begin{figure}[b]
	\centering
	\begin{tikzpicture}
	\begin{axis}[width=\linewidth, height=5.5cm, ymajorgrids=true, xmajorgrids= true, xmin=10, xmax=500000, xmode=log, log basis x=10, ymin=0, ymax=500, ytick distance=50, xlabel={Message Size [Byte]}, ylabel={Round-Trip Latency [ms]}, legend cell align={left}, legend style={at={(.1,.75)},anchor=center,nodes={scale=0.8, transform shape}}]
		
		\addplot[name path=min, draw=none, forget plot] table[x=size, y=min, col sep=semicolon] {plot/connect/tm_sequential_stats.csv};
		
		\addplot[name path=max, draw=none, forget plot] table[x=size, y=max, col sep=semicolon] {plot/connect/tm_sequential_stats.csv};
		
		\addplot[color=gnome_green, opacity=0.3, forget plot] fill between[of=min and max];
		
		\addplot[line width=1pt, dotted, color=gnome_red, forget plot] table[x=size, y=avg, col sep=semicolon] {plot/dds-router/tm_sequential_stats.csv};
		
		\addplot[only marks, mark=*, mark size=1.5, color=gnome_red] table[x=size, y=avg, col sep=semicolon] {plot/dds-router/tm_sequential_stats.csv};
		
		\addplot[line width=1pt, dotted, color=gnome_purple, forget plot] table[x=size, y=avg, col sep=semicolon] {plot/rosbridge/tm_sequential_stats.csv};
		
		\addplot[only marks, mark=*, mark size=1.5, color=gnome_purple] table[x=size, y=avg, col sep=semicolon] {plot/rosbridge/tm_sequential_stats.csv};
		
		\addplot[line width=1pt, dotted, color=gnome_orange, forget plot] table[x=size, y=avg, col sep=semicolon] {plot/zenoh/tm_sequential_stats.csv};
		
		\addplot[only marks, mark=*, mark size=1.5, color=gnome_orange] table[x=size, y=avg, col sep=semicolon] {plot/zenoh/tm_sequential_stats.csv};
		
		\addplot[line width=1pt, dotted, color=gnome_blue, forget plot] table[x=size, y=avg, col sep=semicolon] {plot/connect/tm_sequential_stats.csv};
		
		\addplot[only marks, mark=*, mark size=1.5, color=gnome_blue] table[x=size, y=avg, col sep=semicolon] {plot/connect/tm_sequential_stats.csv};
	
		\addlegendentry{DDS Router}
		\addlegendentry{rosbridge}
		\addlegendentry{Eclipse Zenoh}
		\addlegendentry{ROS2 Connect}
		
	\end{axis}
\end{tikzpicture}
	\caption{\label{plot:seq_connect}Average round-trip latency over a WAN for increasing message sizes. Each data
	point represents the mean of 1000 measurements. Large markers denote measured data points, while small markers indicate 
	interpolated trends. The shaded region indicates the observed latency range for ROS2 Connect.}
\end{figure}

\begin{figure}[b]
	\centering
	\begin{tikzpicture}
	\begin{axis}[width=\linewidth, height=5.5cm, ymajorgrids=true, xmajorgrids= true, xmin=1, xmax=10, ymin=30, ymax=500, ytick distance=50, xtick distance=1, xtick={1, 2, 3, 4, 5, 6, 7, 8, 9, 10}, xlabel={Number of Parallel Topics}, ylabel={Round-Trip Latency [ms]}, legend cell align={left}, legend style={at={(.07,.8)},anchor=center,nodes={scale=0.8, transform shape}}]
		
		\addplot[line width=1pt, dotted, forget plot, color=gnome_purple] table[x=size, y=avg, col sep=semicolon]{plot/connect/tm_parallel_12B_stats.csv};
		
		\addplot[only marks, mark=*, mark size=1.5, color=gnome_purple] table[x=size, y=avg, col sep=semicolon] {plot/connect/tm_parallel_12B_stats.csv};
		
		\addplot[line width=1pt, dotted, forget plot, color=gnome_orange] table[x=size, y=avg, col sep=semicolon]{plot/connect/tm_parallel_10000B_stats.csv};
		
		\addplot[only marks, mark=*, mark size=1.5, color=gnome_orange] table[x=size, y=avg, col sep=semicolon] {plot/connect/tm_parallel_10000B_stats.csv};
		
		\addplot[line width=1pt, dotted, forget plot, color=gnome_blue] table[x=size, y=avg, col sep=semicolon]{plot/connect/tm_parallel_100000B_stats.csv};
		
		\addplot[only marks, mark=*, mark size=1.5, color=gnome_blue] table[x=size, y=avg, col sep=semicolon] {plot/connect/tm_parallel_100000B_stats.csv};
		
		\addlegendentry{12 B}
		\addlegendentry{10 kB}
		\addlegendentry{100 kB}
		
	\end{axis}
\end{tikzpicture}
	\caption{\label{plot:par_connect}Average round-trip latency of ROS2 Connect under concurrent load for increasing numbers of parallel topics.
	Each data point represents the mean of 1000 measurements.}
\end{figure}
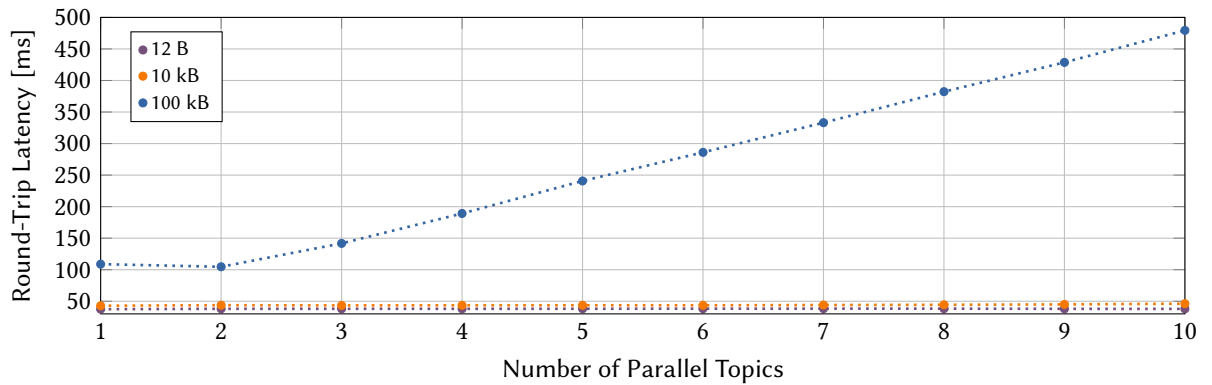

\noindent
Figure \ref{plot:seq_connect} shows the average round-trip latency for increasing message sizes under sequential transmission. Across
all tested configurations, ROS2 Connect consistently achieves the lowest latency compared to DDS Router, rosbridge and Zenoh.
For small message sizes, all four exhibit relatively stable transmission times; however, significant differences emerge as message 
sizes increase.
ROS2 Connect maintains near-constant latency growth over a wide range of payload sizes, indicating efficient and predictable
communication behavior. In contrast, DDS Router shows earlier performance degradation, with latency increasing more rapidly for
larger messages. Furthermore, DDS Router was unable to reliably transmit very large payloads beyond certain sizes in the evaluated
configuration. The rosbridge interface exhibits the highest latency through all measurements due to the overhead introduced by
JSON serialization. Eclipse Zenoh exhibited overall latencies close to those of ROS2 Connect, differing by approximately
1ms for smaller and up to 16ms for larger payloads, with ROS2 Connect consistently achieving lower latency.
In addition to achieving the lowest average latency, ROS2 Connect demonstrates substantially lower variability in round-trip times
compared to DDS Router and rosbridge, while Zenoh exhibited comparable variability.
The narrow latency range observed across most message sizes indicates stable communication behavior, which is particularly important 
for teleoperation scenarios requiring predictable timing characteristics.
Overall, the results confirm that ROS2 Connect provides both improved transmission efficiency and higher latency stability 
compared to existing ROS2 over WAN solutions.
Figure \ref{plot:par_connect} shows the average round-trip latency of ROS2 Connect under concurrent load for increasing number of 
parallel topics.
For small message sizes, latency remains nearly constant even when up to ten parallel topics are transmitted simultaneously, 
indicating that ROS2 Connect efficiently handles concurrent communication with minimal overhead.
For larger message sizes, latency increases approximately linearly with the number of parallel topics. This behavior reflects both 
the limitations of available network bandwidth and the characteristics of WebSocket transport, which relies on a single ordered TCP 
stream and therefore transmits messages sequentially. As a result, large payloads can temporarily occupy the communication channel, 
increasing latency for concurrent transmissions.
Despite this effect, communication remains stable and predictable across all tested configurations. These results demonstrate that 
ROS2 Connect scales effectively under concurrent load, with small control and sensor messages remaining largely unaffected while 
larger messages primarily reflect transport-layer constraints inherent to wide-area communication.

\noindent
Due to space constraints, only representative results are presented here. A more extensive evaluation, including additional 
performance analyses and implementation details, is provided in the author's accompanying master's thesis \cite{schott}.

\section{Conclusions}

This work presented ROS2 Connect, a WebSocket-based communication framework designed to enable secure and reliable ROS2 interactions
across WANs. By decoupling ROS2 communication from multicast-dependent DDS discovery, the proposed approach enables
transparent transmissions of topics, services, actions, and system infrastructure data without requiring network-layer modifications
or vendor-specific configurations.
Experimental evaluation demonstrated that ROS2 Connect achieves lower latency, higher stability, and better scalability under 
the evaluated concurrent load compared to existing ROS2 over WAN solutions, including DDS Router, rosbridge, and Zenoh. The results 
further showed that communication performance remains predictable across a wide range of message sizes and parallel workloads, making 
the framework well suited for teleoperation scenarios.
Further work will focus on extending generic support for services and actions, evaluating in real robotic deployments with full application stacks, and exploring adaptive transport optimizations under varying network conditions.

\begin{acknowledgments}
	The authors gratefully acknowledge the Virtual University of Bavaria (VHB 24-II-02-13-Nue1) for its continuous support since 2005 and
  for the grant that partially funded the implementation presented in this work.
\end{acknowledgments}

\section*{Declaration on Generative AI}

During the preparation of this work, the authors used DeepL Write and ChatGPT 5.2 in order to: Grammar and spelling check, 
Paraphrase and reword.
After using these tools/services, the authors reviewed and edited the content as needed and takes full responsibility for the publication’s content. 

\bibliography{ros2-connect}

@mastersthesis{schott,
	Author = {Daniel Schott},
	Title = {{ROS2 Over Wide-Area-Networks: Teleoperation of Mobile Robots in an E-Learning Environment}},
	School = {Julius-Maximilians-University},
	Year = {2025},
	Address = {Würzburg, Bavaria, Germany},
	Month = {September},
	Url = {https://robotik.informatik.uni-wuerzburg.de/telematics/download/MSc_Daniel_Schott.pdf}
}

@article{ros2,
	Title = {{Robot Operating System 2: Design, architecture, and uses in the wild}},
	Author = {Steven Macenski and Tully Foote and Brian Gerkey and Chris Lalancette and William Woodall},
	Journal = {Science Robotics},
	Volume = {7},
	Number = {66},
	Pages = {eabm6074},
	Year = {2022},
	Doi = {10.1126/scirobotics.abm6074}
}

@techreport{dds,
	Title = {{Data Distribution Service (DDS), Version 1.4}},
	Author = {{Object Management Group}},
	Institution = {{Object Management Group}},
	Address = {9C Medway Road, PMB 274 Milford, MA 01757 USA},
	Year = {2015},
	Month = {April},
	Url = {https://www.omg.org/spec/DDS/1.4/PDF}
}

@techreport{rtps,
	Title = {{The Real-time Publish-Subscribe Protocol (RTPS) DDS Interoperability Wire Protocol Specification, Version 2.2}},
	Author = {{Object Management Group}},
	Institution = {{Object Management Group}},
	Address = {9C Medway Road, Milford, MA 01757 USA},
	Year = {2014},
	Month = {September},
	Url = {https://www.omg.org/spec/DDSI-RTPS/2.2/PDF}
}

@article{multicast,
	Author={Christophe Diot and Brian Neil Levine and Bryan Lyles and Hassan Kassem and Doug Balensiefen},
	Journal={IEEE Network}, 
	Title={Deployment issues for the IP multicast service and architecture}, 
	Year={2000},
	Volume={14},
	Number={1},
	Pages={78-88},
	Doi={10.1109/65.819174}
}

@techreport{fastdds,
	Title = {{Fast DDS Documentation, Release 3.4.2}},
	Author = {{eProsima}},
	Institution = {{eProsima}},
	Address = {Plaza de la Encina 10-11 Nucleo 4 2ª Planta, 28760 Tres Cantos, Madrid, Spanien},
	Year = {2026},
	Month = {January},
	Url = {https://fast-dds.docs.eprosima.com/_/downloads/en/v3.4.2/pdf/}
}

@misc{cyclonedds,
	Title = {{Cyclone DDS, Version 0.10.5}},
	Author = {{Eclipse Foundation}},
	Url = {https://cyclonedds.io/docs/cyclonedds/0.10.5/},
	Year = {2024},
	Month = {May}
}

@techreport{ddsrouter,
	Title = {{DDS Router Documentation, Release 3.4.0}},
	Author = {{eProsima}},
	Institution = {{eProsima}},
	Address = {Plaza de la Encina 10-11 Nucleo 4 2ª Planta, 28760 Tres Cantos, Madrid, Spanien},
	Year = {2025},
	Month = {November},
	Url = {https://eprosima-dds-router.readthedocs.io/_/downloads/en/v3.4.0/pdf/}
}

@misc{rosbridge,
	Title = {{rosbridge\_suite}},
	Author = {{Robot Web Tools}},
	Url = {https://github.com/RobotWebTools/rosbridge\_suite},
	Year = {2026},
	Note = {last accessed: February 16, 2026}
}

@misc{zenoh,
	Title = {{Zenoh}},
	Author = {{Eclipse Foundation}},
	Url = {https://zenoh.io/},
	Year = {2026},
	Note = {last accessed: February 16, 2026}
}

@article{laks,
	Title = {{Design and Development of a Robotic Teleoperation System using Duplex WebSockets suitable for Variable Bandwidth Networks}},
	Author = {Lakshminarasimhan Srinivasan and Julian Scharnagl and Zhihao Xu and Nicolas Faerber and Dinesh Kumar Babu and Klaus Schilling},
	Journal = {IFAC Proceedings Volumes},
	Volume = {46},
	Number = {29},
	Pages = {57-61},
	Year = {2013},
	Note = {3rd IFAC Symposium on Telematics Applications}	
}

\appendix

\section{Source Code}

ROS2 Connect is released under the Mozilla Public License Version 2.0
and is available at:\\
\url{https://github.com/JMUWRobotics/ROS2-Connect}

\end{document}